\documentclass{article}
\usepackage{amsmath,graphicx,mlspconf}
\usepackage{xcolor}

\usepackage{amsfonts}       
\usepackage{booktabs}

\def\cN{\mathcal{N}}

\title{HALFNET: RANDOMIZED NEURAL NETWORKS WITH LEARNED SUBSPACE GEOMETRY}

\name{Ethem Alpayd{\i}n}
\address{Independent Researcher\\
Istanbul, Turkey\\
ethem.alpaydin@gmail.com
}

\begin{document}

\maketitle

\begin{abstract}
Many researchers investigated neural networks with some of their weights fixed to values randomly drawn from a given distribution, e.g., $\cN(0, I)$. Our proposed HalfNet draws random weights from $\cN(0, \Sigma)$, where $\Sigma$, which defines the geometry of the distribution, has a low-rank factorization that we learn from data. Experiments on MNIST and CIFAR-10 demonstrate that HalfNet can match the performance of fully trained multilayer perceptrons while using substantially fewer parameters. Spectral analysis  indicates that much of the predictive power of neural networks lies in the geometry of their weight space rather than in the precise values of individual parameters and we observe that accuracy scales smoothly with rank. HalfNet is not a neural architecture trick for low-rank structure; it implements a data-dependent random embedding that can also be interpreted through supervised metric learning, or random-feature and kernel perspectives.
\end{abstract}
\begin{keywords}
Randomized neural networks, low-rank learning, metric learning, random features, kernels
\end{keywords}

\section{Introduction}

As an alternative to the usual approach of neural networks with all their weights trained on data, one can fix some of the weights to random values  \cite{rahimi07, huang12, cao17, scar17, gall20, alp25}. This leads to a simpler model and a simpler optimization problem. The Johnson-Lindenstrauss lemma states that a high-dimensional space can be embedded into a space of much lower dimension using random projections in such a way that distances between the points are nearly preserved.

In the type of layer that we propose here, we learn the distribution from which the random weights are to be drawn. That is, rather than learning every weight individually, our proposed {\em HalfNet} learns the distribution from which the random projection vectors are sampled. Our experimental results show that this improves accuracy because learning the distribution allows a better match of the model with the structure embedded in the data, without needing to learn each of the weights separately. 

In particular, we assume that the weights of a layer are drawn from $\cN(0, \Sigma)$, and we use a low-rank parameterization of $\Sigma$, which we learn. That is, the projection vectors remain random while the geometry of their distribution is learned. By changing the rank parameter, we get a set of models of increasing complexity interpolating between fully random units and fully trainable networks. 

In summary, our contributions are as follows:
\begin{enumerate}
\item We introduce HalfNet, which learns the covariance structure of random projections rather than the individual weights.
\item Accuracy scales smoothly with rank, indicating that much of the predictive power of trained layers is captured by the low-dimensional geometric structure.
\item We show that the framework can be extended to (i) multiple layers, (ii) convolutional front ends, and (ii) binary weights.
\item HalfNet admits probabilistic, metric-learning, and kernel interpretations.
\end{enumerate}

This paper is organized as follows: We present our model in Section~\ref{sec-model} and our experimental results on MNIST in Section~\ref{sec-mnist}. In Section~\ref{sec-bin}, we show how our proposed model can also be adapted to learn binary weights. We give our experimental results on CIFAR-10 in Section~\ref{sec-cifar}. Section~\ref{sec-rel} discusses related work and the relationship to metric learning and kernels. We conclude in Section~\ref{sec-conc}.

\section{Model Formulation}
\label{sec-model}

We have hidden unit $h$ with $x, w_h \in \mathbb{R}^d$ and $y_h, w_{h0}\in\mathbb{R}$
\begin{equation}
y_h = g(w_h^\top x + w_{h0} )	
\label{eq-full}
\end{equation}

\noindent where typically, $w_h, w_{h0}$, are trained on a dataset.

The weights may be drawn randomly, then kept fixed and not updated during learning:
\begin{equation}
y_h = g(r_h^\top x ), \quad r_h \sim \mathcal{N}_d(0, I). 	
\label{eq-random}
\end{equation}

With HalfNet, our aim is to learn the distribution of $r_h$ from data. We define a low-rank factor $B \in \mathbb{R}^{d\times k}$ with $k < d$ and a diagonal scaling $\sigma \in \mathbb{R}^d$ and define
\begin{align}
r_h 	&= B z_h + \mathrm{diag}(\sigma)\,\epsilon_h  \mbox{\ where\ }
	z_h \sim \mathcal{N}_k(0,I), \epsilon_h \sim \mathcal{N}_d(0,I)
\label{eq-lowrank}
\end{align}

\noindent and we get $r_h \sim \mathcal{N}_d(0, \Sigma)$ with $\Sigma= BB^\top + \mathrm{diag}(\sigma^2)$. $(z_h,\epsilon_h)$ are sampled once and fixed; $B, \sigma$ are learned. 

Once $r_h$ are generated, they are further adjusted \cite{alp25}:
\begin{align}
y_h	&= g(u_h (r_h ^\top x) + u_{h0})
\end{align}

\noindent where $u_h$ and $u_{h0}$ are also learned. These additional parameters (two per unit) scale and translate the randomly placed hyperplanes (defined by $r_h$).

To summarize, instead of drawing $d$-dimensional random weights from $\mathcal{N}_d(0, I)$, we draw $k < d$-dimensional random vectors $z_h \sim \mathcal{N}_k(0, I)$ and expand them into $d$ dimensions using the matrix $B$. We also draw $\epsilon_h \sim \mathcal{N}_d(0, I)$ and scale it element-wise by $\sigma$. These two components are then added to form the $d$-dimensional vector $r_h$. The random vectors $z_h$ and $\epsilon_h$ are different for each hidden unit, while $B$ and $\sigma$ are shared across all units in the same layer.  Finally, the parameters $u_h$ and $u_{h0}$ are specific to each hidden unit and allow further adaptation at the unit level.

The rank $k$ of $B$ is a hyperparameter that is selected by cross-validation, and varying that we get HalfNet variants of different complexities allowing us to match the bias-variance tradeoff. For a fully-connected layer with input dimension $d$ and $m$ units, in the standard formulation, there are $dm + m$ weights; with the HalfNet parameterization, there are  $dk + d + 2m$ trainable parameters. 

\section{Experiments on MNIST}
\label{sec-mnist}

We use the MNIST dataset which has $28\times 28=784$ inputs and ten classes, with 60,000 training and 10,000 test instances. We use ReLU as the activation function at the hidden units and softmax at the output; training is done to minimize the cross-entropy using Adam. The weights to the outputs are trained as usual; only the hidden units are half units.

Each configuration is run five times starting from different random seeds and we plot mean values with one standard deviation error bars vs. the number of modifiable parameters. We compare the HalfNet variants with the linear perceptron (LP) and the multilayer perceptron (MLP) with all their weights trained as usual.

\subsection{Results With One Hidden Layer}

\begin{figure}[htbp]
\begin{center}
\includegraphics[width=8cm]{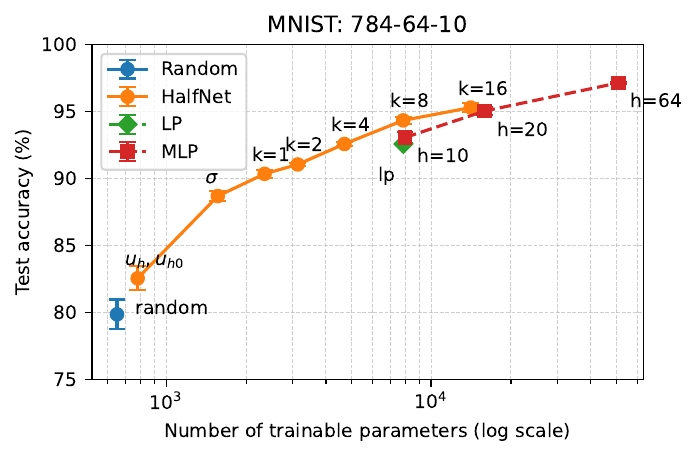} 
\end{center}
\caption{The number of parameters vs test accuracy on MNIST with 784--64--10.}
\label{fig_mnist_sma_noconv}
\end{figure}

Our first architecture has one hidden layer with 64 hidden units, 784--64--10, and the results are plotted in Figure~\ref{fig_mnist_sma_noconv}.
When all the weights are random, we have a convex problem and only the second layer to train and this already achieves good accuracy. 

When we add unitwise parameters $u_h, u_{h0}$ on top of random weights, we see that we get an improvement. If we add $\sigma$, also unitwise but shared by all units, we get a big improvement; this shows that (as expected) all inputs (pixels) are not equally important. Then when we add the low-rank $B$, accuracy increases as the rank $k$ is increased. On a dataset like MNIST where there is strong dependency between the input pixels, taking the correlation between the inputs into account naturally leads to improvement. We see that with $k=4$ we get as high accuracy as LP with almost half the parameters; with $k=16$, we match the accuracy of MLP with a comparable number of parameters. 

\subsection{Spectral Analysis}

\begin{figure}[htbp]
\begin{center}
\begin{tabular}{cc}
\includegraphics[width=4cm]{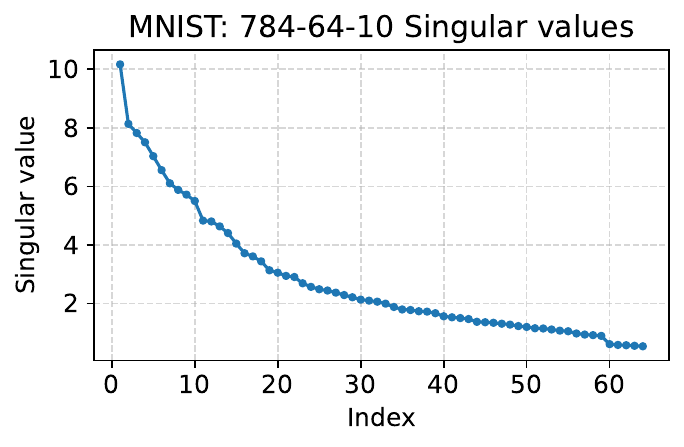} &
	\includegraphics[width=4cm]{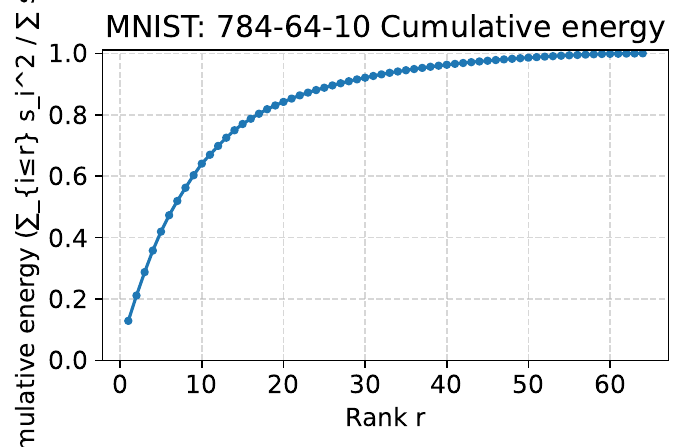} 
\end{tabular}
\caption{Singular values and the total energy explained on MNIST.}
\label{fig-mnist-sing}
\end{center}
\end{figure}

To get a better understanding of what our proposed model is doing, we take the $784\times 64$ matrix of first-layer weights of a fully trained 64-unit MLP and compute its SVD. On MNIST, the singular values and the total energy explained by the leading $k$ directions are plotted in Figure~\ref{fig-mnist-sing}.

We see that the first-layer weight matrix exhibits a strongly anisotropic spectrum: 60\% of its spectral energy is in the leading nine singular directions and 80\% in the first 17 directions. The rapid decay of the singular values indicates that the first-layer weight matrix learned by a fully trained MLP lies largely within a low-dimensional subspace. The low-rank HalfNet variant explicitly parameterizes such a subspace through the smaller matrix $B$. 

This explains why small values of $k$ can recover much of the predictive performance. With $k=8$ and $k=16$, HalfNet achieves 94.34\% and 95.30\% test accuracy; the long spectral tail contributes additional, but relatively smaller performance gains: the full 64-unit MLP achieves 97.14\% test accuracy.

Because the weights are constrained by the covariance structure, HalfNet behaves like a {\em geometric regularizer}. But note that HalfNet is not just a low-rank neural layer. In a low-rank layer, the weight matrix is explicitly parameterized as $W = U B$ and learned deterministically. In contrast, HalfNet learns the covariance of a distribution from which projection vectors are sampled. The resulting features are therefore stochastic random features whose geometry is controlled by the learned covariance matrix. 

\subsection{Results With A Convolutional Frontend}

A feature-extracting frontend can be placed in front of HalfNet and its parameters can be trained concurrently. We include a single convolutional layer with 16 filters of size $5\times5$, followed by max pooling ($2\times2$), which after flattening produces a feature vector of 2,304 dimensions. This is then fed to the HalfNet or any other classifier. 

This convolutional layer itself adds only 416 parameters, but because the input dimensionality increases from 784 to 2,304, it significantly increases the total number of parameters with fully-connected LP and MLP, but not HalfNet because its complexity does not depend on the fan-in but on the effective rank, which is still small.

\begin{figure}[htbp]
\begin{center}
\includegraphics[width=8cm]{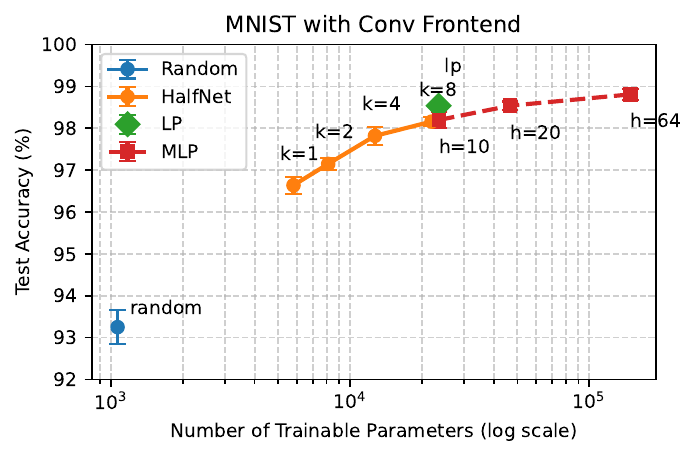} 
\caption{Results on MNIST with convolutional frontend: $(28\times28)$--2304--64--10.}
\label{fig_mnist_sma_conv}
\end{center}
\end{figure}

Our results on MNIST are plotted in Figure~\ref{fig_mnist_sma_conv}. All models, including HalfNet profit from the frontend; the problem becomes almost linearly separable and MLP leads to no significant improvement over LP. We see again that HalfNet variants generate a spectrum of models with increasing complexity leading to incrementally better accuracy. Even with $k=1$, we get quite high accuracy.

\subsection{Results With Two Hidden Layers}

We also test the model with two hidden layers, both having 64 hidden units: 784--64--64--10. So we have two layers of half-units, and the parameters ($B, \sigma$) of these two hidden layers are kept separate.

\begin{figure}[htbp]
\begin{center}
\includegraphics[width=8cm]{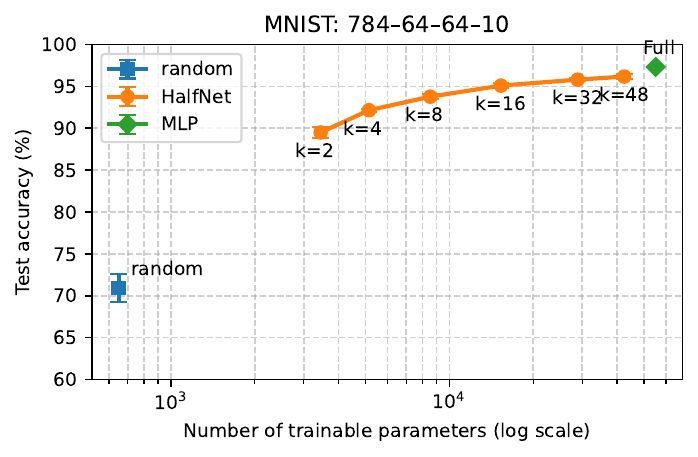} 
\caption{The number of parameters vs test accuracy on MNIST with 784--64--64--10.}
\label{fig_mnist_lar_noconv}
\end{center}
\end{figure}

The results on MNIST are plotted in Figure~\ref{fig_mnist_lar_noconv}. We see again that HalfNet achieves accuracies higher than what is achieved by purely random units, and its accuracy increases as $k$ goes up. This experiment shows that one can use a HalfNet layer anywhere in a large network, and we can have multiple HalfNet layers in the same network.

\section{Half Layer Networks with Binary Weights}
\label{sec-bin}

The model that we discuss above can be adapted to generate binary ($\pm$1) weights, which are attractive because they can lead to major savings in terms of memory and computation. We can generate binary hyperplanes by placing the sign function after Equation~\ref{eq-lowrank}. This allows us to map $\cN_k(0,I)$ to a $d$-dimensional multivariate binary (Rademacher) distribution where the resulting binary dimensions become dependent through the covariance structure $\Sigma$. 

Actually, the final distribution corresponds to the sign of the learned $\mathcal{N}_d(0,\Sigma)$.  Although the sign operation discards magnitude information, it preserves angular relationships between projections, so the correlation structure induced by $\Sigma$ is retained in the binary features. The sign function is not differentiable, but following standard practice, we employ a straight-through estimator (STE) during back-propagation. 

Such constructions are often referred to as Gaussian-copula Bernoulli models. A naive random network with binary weights would sample them from a multivariate Bernoulli distribution with independent dimensions; our approach allows us to learn the dependencies also in this case of binary weights. As in the continuous case, increasing the latent dimension $k$ allows representing more complex correlation patterns among the binary hyperplanes.

\begin{figure}[h]
\begin{center}
\includegraphics[width=8cm]{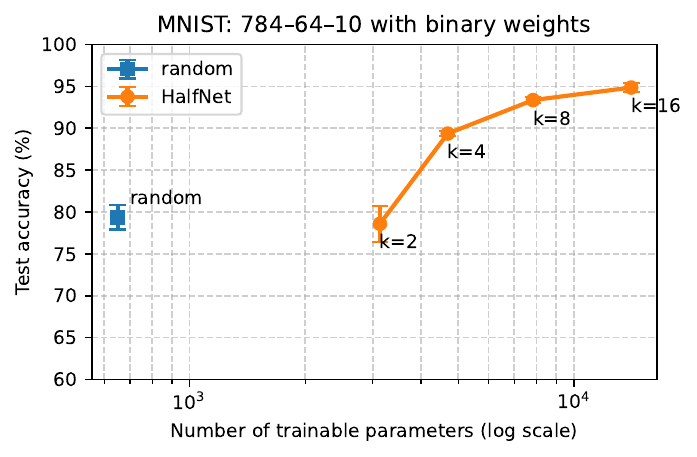} 
\caption{The number of parameters vs test accuracy on MNIST with binary weights.}
\label{fig_mnist_sma_bin}
\end{center}
\end{figure}

Again using the network with one hidden layer but with binary weights in the half-layer, we get the results plotted in Figure~\ref{fig_mnist_sma_bin} on MNIST. The random version that assumes independent Bernoullis is not bad. With HalfNet, $k=2$ seems too low for this binary case with its limited precision. This indicates that at very small latent dimensionality, binarization can collapse feature diversity leading to a drop in accuracy. But with $k \ge 4$, the model gets quite accurate: With $k=16$, with continuous weights we get $95.30\%$ on the test set; with binary weights, we get $94.87\%$. 

\section{Experiments on CIFAR-10}
\label{sec-cifar}

To evaluate the applicability of HalfNet beyond relatively simple grayscale datasets such as MNIST, we conduct additional experiments on the CIFAR-10 dataset. CIFAR-10 consists of 60,000 color images of size $32 \times 32$ belonging to ten classes, with 50,000 training and 10,000 test instances. Compared to MNIST, CIFAR-10 presents a significantly more challenging task due to higher input dimensionality, color channels, and more complex spatial structure.

On MNIST, we observed that HalfNet is particularly effective in fully-connected layers with large fan-in, where the learned low-dimensional geometry captures most of the discriminative structure. In contrast, convolutional layers operate on small spatial neighborhoods (e.g., $5 \times 5$ filters), and the benefit of low-rank structure is less obvious a priori. Nevertheless, convolutional filters within a layer are often highly correlated, suggesting that they may also lie in a low-dimensional subspace. This motivates extending HalfNet to convolutional layers by assuming that each filter is drawn from a learned distribution with shared covariance structure.

Concretely, for a convolutional layer with filters $w_h \in \mathbb{R}^{c \times p \times p}$ (flattened as vectors in $\mathbb{R}^d$), we generate filters as
\[
w_h = B z_h + \mathrm{diag}(\sigma)\,\epsilon_h
\]

\noindent where $B \in \mathbb{R}^{d \times k}$ and $\sigma \in \mathbb{R}^d$ are learned and shared across filters, and $(z_h, \epsilon_h)$ are fixed random variables. This introduces a form of structural sharing across filters: rather than learning each one independently, the model learns a low-dimensional family of filters.

To isolate the effect of this parameterization, we use a simple architecture with a single convolutional layer followed by a fully-connected classifier. The convolutional layer has 32 filters of size $3 \times 5 \times 5$, followed by $2 \times 2$ pooling, resulting in a feature vector of dimension 8,192; this layer has 2,432 trainable parameters. The fully-connected hidden layer has 256 units, with $(8192 + 1) \times 256 = 2,097,408$ parameters, and the final layer maps to the ten output units with $(256 + 1) \times 10 = 2,560$ parameters. We keep the output layer as is and apply the HalfNet parameterization to the convolutional and/or the first fully-connected layer.

While the convolutional layer has only 2,432 parameters, the first fully-connected hidden layer has over 2 million parameters. Thus, more than 99\% of the model capacity is concentrated in a single dense layer, and this is what can profit most from HalfNet. 

We compare three configurations:
\begin{itemize}
\item Full--Full: both convolutional and fully-connected layers are fully trainable.
\item Full--Half: the convolutional layer is fully trainable, while the fully-connected layer is a HalfNet layer.
\item Half--Half: both the convolutional and fully-connected layers use HalfNet parameterization.
\end{itemize}

Training is performed using Adam with cross-entropy loss; all models are trained for 50 epochs and we calculate the mean and standard deviation over five runs.

\begin{figure}[h]
\begin{center}
\includegraphics[width=8cm]{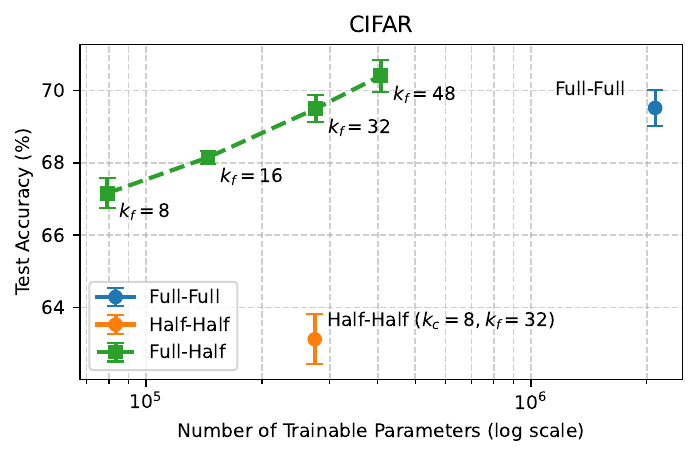} 
\caption{The number of parameters vs test accuracy of the different models on CIFAR-10.}
\label{fig_cifar}
\end{center}
\end{figure}

We see in Figure~\ref{fig_cifar} that with the Full--Half configuration, increasing the rank $k_f$ of the fully-connected layer leads to rapid gains in accuracy even at small values. With $k_f=32$, Full--Half matches the test accuracy of Full--Full while using approximately $8\times$ fewer parameters. This confirms our previous observations on MNIST that much of the predictive capacity of a dense layer can be captured by learning a low-dimensional covariance structure rather than individual weights. More notably, for $k_f=48$, we slightly exceed the performance of the fully trained baseline while still using significantly fewer parameters.

When HalfNet is also applied to the convolutional layer too (Half--Half) with $k_c=8, k_f=32$, performance decreases relative to Full--Half, and training becomes slower. This shows that the assumption of low-dimensional structure may be restrictive; of course, the accuracy of Half--Half can be increased by using higher $k_c$, but given that convolutional layers already benefit from strong structural constraints such as locality and weight sharing, their parameter count is relatively small, and additional low-rank regularization appears less beneficial.

\section{Relationship to Low-Rank Models, Metric Learning, and Kernels}
\label{sec-rel}

A large body of work has explored low-rank structure in neural networks, both as an implicit bias and as an explicit parameter reduction technique. Empirical and theoretical studies have shown that deep networks often learn representations that are effectively low-dimensional \cite{huh2021lowrank, oymak2019generalization}. In addition, low-rank factorizations and tensor decompositions have been widely used to compress neural networks by approximating weight matrices after or during training \cite{denil2013predicting, jaderberg2014speeding}. 

In the classical (deterministic) low-rank formulation, we parameterize the $d\times m$ weight matrix as $W=UV$, with $U\in\mathbb{R}^{d\times k}$ and $V\in\mathbb{R}^{k\times m}$, and both $U$ and $V$ are learned. 

Our approach can be written (ignoring the $\sigma$ and $u_h, u_{h0}$) as $W=BZ$ where $B\equiv U$ and the columns of $Z$ are drawn from $\cN_k(0, I)$. So $Z$ are fixed and only $B$ is learned. $z_h$ are random projections, and we can interpret the role of $B$ as discovering a $k$-dimensional coordinate system in which simple random projections are already useful. Our experiments above show that this can be done with almost no loss of accuracy; that is, instead of learning both factors, we can just fix one to random projections (which also leads to a simpler optimization problem).

This distinction is also reflected in related approaches based on random features and kernel approximations \cite{rahimi07, bach2013nystrom}, where fixed random projections are used to approximate high-dimensional feature mappings. In contrast, HalfNet learns the geometry of the projection distribution itself, allowing the representation to adapt to the data.

From a probabilistic perspective, our formulation is also related to Gaussian and Bayesian views of neural networks \cite{neal1996bayesian, lee2018deep}, where distributions over weights induce structured function spaces. In our case, $\Sigma$ directly encodes the geometry of the weight space and is learned end-to-end.

HalfNet has three complementary interpretations:
\begin{enumerate}
\item \textbf{Random projection view.}
Hidden units compute projections of the form
\[
a = r^\top x
\]
where $r \sim \mathcal{N}_d(0,\Sigma)$. Writing $r = Bz$ with $B \in \mathbb{R}^{d \times k}$ and $z \sim \mathcal{N}_k(0,I)$ gives
\[
a = z^\top B^\top x .
\]
Defining the transformed input $x' = B^\top x$, the computation becomes
\[
a = z^\top x' .
\]
Thus HalfNet can be interpreted as first learning a linear embedding $x' = B^\top x$ that reshapes the geometry of the input space (decreasing dimensionality from $d$ to $k$), followed by fixed random projections (in the $k$-dimensional space).

\item \textbf{Metric-learning view.}
Let the learned covariance be $\Sigma = BB^\top$. The induced quadratic form
\[
\|x\|_{\Sigma}^{2} = x^\top \Sigma x
\]
defines a Mahalanobis metric. The distance between two points becomes
\[
d_{\Sigma}(x,y)^2
=
(x-y)^\top \Sigma (x-y)
=
\|B^\top x - B^\top y\|^2 .
\]
Thus learning $B$ corresponds to learning which directions in the input space are emphasized or suppressed. In this view, HalfNet performs supervised low-rank metric learning.

\item \textbf{Kernel view.}
Each hidden unit defines a random feature
\[
\phi_r(x) = g(r^\top x)
\]
with $r \sim \mathcal{N}_d(0,\Sigma)$ which defines a random kernel \cite{rahimi07}
\[
K_\Sigma(x,y)
=
\mathbb{E}_{r \sim \mathcal{N}(0,\Sigma)}
\big[g(r^\top x) g(r^\top y)\big].
\]
Thus, HalfNet can be viewed as a random feature method in which the distribution of projection vectors is not fixed but learned through $\Sigma$. Learning $\Sigma$ therefore corresponds to learning the kernel implicitly, adapting the geometry of the feature space to the data.
\end{enumerate}

\section{Conclusions}
\label{sec-conc}

We introduced HalfNet, a stochastic parameterization in which random weights are drawn from a distribution whose covariance structure is learned from data. The rank of the covariance factor provides a direct control over model capacity, yielding a smooth trade-off between fixed random features and fully trainable neural networks.

Across MNIST and CIFAR-10 experiments, HalfNet matches fully trained baselines with substantially fewer parameters, especially in high-dimensional fully-connected layers. Spectral analysis supports this behavior by showing that trained weight matrices often concentrate much of their energy in a low-dimensional subspace.

HalfNet is distinct from deterministic low-rank factorization: it does not learn a single low-rank weight matrix, but rather a distribution over projection vectors whose geometry is adapted to the task. This connects the method to random features, metric learning, and kernel learning, while preserving a simple neural-network implementation.

We also showed that (i) a multilayer network can have multiple HalfNet layers, (ii) a HalfNet layer can be placed after a convolutional layer, (iii) the HalfNet approach can also be used to train a convolutional layer, and (iv) HalfNet can also be adapted to generate binary weights.

Our experiments are restricted to image datasets where correlations in the data are pronounced. The benefits of HalfNet may be reduced in settings with weaker correlations, and evaluating the approach on a broader range of datasets remains an important direction for future work. At the same time, HalfNet can be most effective in high fan-in settings with strong correlations, such as residual or attention layers in larger networks. 

Overall, HalfNet demonstrates that learning the geometry of projection distributions can capture much of the predictive power of fully trained neural layers while using substantially fewer parameters.

\section*{Acknowledgments}

The author used ChatGPT (OpenAI) for assistance with code generation, mathematical formatting, and discussion of conceptual interpretations during the preparation of this work.
\bibliographystyle{IEEEbib}
\bibliography{half26}



\newpage
\appendix

\setcounter{figure}{0}
\renewcommand{\thefigure}{A\arabic{figure}}

\setcounter{table}{0}
\renewcommand{\thetable}{A\arabic{table}}

\section{Results on MNIST}

\begin{table}[h]
\caption{Results on MNIST with one hidden layer: 784--64--10.}
\label{tab_mnist_sma_noconv}
\centering
\begin{tabular}{llrrr}
\toprule
Model & Hyperpars	& Params & Train acc & Test acc \\
\midrule
LP		&		& 7,850	& 92.84, 0.02 	& 92.58, 0.09 	\\
\hline
MLP		& $h=10$	& 7,960	& 93.36, 0.42	& 93.06, 0.41 	\\
		& $h=20$	& 15,910	& 95.63, 0.24	& 95.04, 0.31 	\\
		& $h=64$	& 50,890	& 98.36, 0.08	& 97.14, 0.06	\\
\hline
random	&  	& 650	& 79.33, 1.23	&  79.86, 1.11	\\
\hline
HalfNet	& $u_h, u_{h0}	$	& 778	& 82.02, 1.10	& 82.55, 0.90	\\
		& $\sigma$	& 1,562	& 88.67, 0.32	& 88.69, 0.36	\\
		& $k=1$	& 2,346	& 90.38, 0.33	& 90.35, 0.29 \\			
		& $k=2$	& 3,130	& 91.33, 0.18	& 91.05, 0.14 \\
		& $k=4$	& 4,698	& 93.11, 0.13	& 92.58, 0.18 \\
		& $k=8$	& 7,834	& 95.06, 0.23	& 94.34, 0.25 \\
		& $k=16$	& 14,106	& 96.39, 0.26	& 95.30, 0.31 \\
\bottomrule
\end{tabular}
\end{table}

\begin{table}[h]
\centering
\caption{Results on MNIST with convolutional frontend and one fully-connected hidden layer: $(28\times 28)$--2304--64--10.}
\medskip
\begin{tabular}{llrrr}
\toprule
Model 	& Hyperpars	& Params & Train acc & Test acc \\
\midrule
LP		&		& 23,466	& 99.20, 0.05	& 98.54, 0.04 \\
\hline
MLP		& $h=10$	& 23,576	& 98.60, 0.20	& 98.20, 0.19 \\
		& $h=20$	& 46,726	& 99.12, 0.15	& 98.54, 0.10 \\
		& $h=64$	& 148,586	& 99.62, 0.08	& 98.81, 0.13 \\
\hline
random		&  	& 1,066	& 93.08, 0.25	&  93.25, 0.40	\\
\hline
HalfNet	& $k=1$	& 5,802	& 96.98, 0.13	& 96.64, 0.20 \\			
		& $k=2$	& 8,106	& 97.70, 0.14	& 97.15, 0.15 \\
		& $k=4$	& 12,714	& 98.46, 0.13	& 97.82, 0.22 \\ 
		& $k=8$	& 21,930	& 98.90, 0.06	& 98.16, 0.11 \\
\bottomrule
\end{tabular}\label{tab_mnist_sma_conv}
\end{table}

\begin{table}[h]
\centering
\caption{Results on MNIST with two hidden layers: 784--64--64--10.}
\medskip
\begin{tabular}{llrrr}
\toprule
Model 	& Hyperpars	& Params	& Train acc 	& Test acc \\
\midrule
random	& 			& 650		& 60.97, 1.88	& 70.92, 1.71 \\
\hline
HalfNet 	& $k=2$	& 3,450	& 89.93, 0.37	& 89.50, 0.68 \\
HalfNet	& $k=4$	& 5,146	& 92.74, 0.21	& 92.16, 0.18 \\
HalfNet	& $k=8$	& 8,538	& 94.66, 0.17	& 93.78, 0.29 \\
HalfNet 	& $k=16$	& 15,322	& 96.35, 0.25	& 95.10, 0.27 \\
HalfNet	& $k=32$	& 28,890	& 97.67, 0.17	& 95.81, 0.18 \\
HalfNet 	& $k=48$	& 42,458	& 98.15, 0.16	& 96.18, 0.26 \\
\hline
MLP 		&		& 55,050	& 98.71, 0.25	& 97.33, 0.15 \\
\bottomrule
\end{tabular}\label{tab_mnist_lar_noconv}
\end{table}

\begin{table}[h]
\centering
\caption{Results on MNIST with one hidden layer (784--64--10) and  binary half-layer weights.}
\medskip
\begin{tabular}{llrrr}
\toprule
Model 		& Hyperpars	& Params	& Train acc 	& Test acc \\
\midrule
random		&			& 650	& 78.42, 1.52	& 79.35, 1.44 \\
\hline
HalfNet		& $k=2$	& 3,130	& 78.13, 2.27	& 78.59, 2.12 \\
			& $k=4$	& 4,698	& 89.67, 0.32	& 89.37, 0.34 \\
			& $k=8$	& 7,834	& 93.96, 0.28	& 93.37, 0.37 \\
			& $k=16$	& 14,106	& 95.38, 0.42	& 94.87, 0.55 \\
\bottomrule
\end{tabular}\label{tab_mnist_bin}
\end{table}

\newpage
\section{Results on Fashion-MNIST}

We also have results on Fashion-MNIST which is a dataset that has the same properties as MNIST. We use the same architectures also on this dataset, but due to page limitations and because the findings are in parallel with what we get on MNIST, those results are given here.

\begin{table}[htbp]
\centering
\caption{Results on Fashion-MNIST with one hidden layer: 784--64--10. }
\medskip
\begin{tabular}{llrrr}
\toprule
Model 	& Hyperpars	& Params		& Train acc 	& Test acc \\
\midrule
LP		&		& 7,850		& 86.35, 0.13 	& 84.31, 0.13 \\
\hline
MLP		& $h=10$	& 7,910		& 85.20, 0.53	& 83.51, 0.35 \\
		& $h=20$	& 15,910		& 87.29, 0.55	& 85.28, 0.63 \\
		& $h=64$	& 50,890		& 89.47, 0.21	& 87.16, 0.25 \\
\hline
		& random 	&  650		& 76.60, 0.71	& 75.51, 0.76	\\
\hline
HalfNet	& $k=1$	& 2,346	& 83.49, 0.33	& 81.95, 0.24 \\			
		& $k=2$	& 3,130	& 84.27, 0.33	& 82.85, 0.11 \\
		& $k=4$	& 4,695	& 85.53, 0.29	& 83.71, 0.55 \\
		& $k=8$	& 7,834	& 87.00, 0.25	& 84.90, 0.37 \\
		& $k=16$	& 14,106	& 87.96, 0.36	& 85.70, 0.36\\
\bottomrule
\end{tabular}\label{tab_fmnist_sma_noconv}
\end{table}

\begin{table}[htbp]
\centering
\caption{Results on Fashion-MNIST with convolutional frontend and one fully-connected hidden layer: $28\times 28$--2304--64--10.}
\medskip
\begin{tabular}{llrrr}
\toprule
Model 	& Hyperpars	& Params & Train acc & Test acc \\
\midrule
LP		&		& 23,466	& 91.39, 0.18	& 89.46, 0.29 \\
\hline
MLP		& $h=10$	& 23,576	& 89.10, 0.65	& 87.96, 0.62 \\
		& $h=20$	& 46,726	& 91.12, 0.51	& 89.31, 0.27 \\
		& $h=64$	& 148,586	& 93.08, 0.40	& 90.41, 0.25 \\
\hline
		& random 	& 1,066	& 82.85, 0.32	&  81.66, 0.57	\\
\hline
HalfNet	& $k=1$	& 5,802	& 87.76, 0.38	& 86.36, 0.47 \\
		& $k=2$	& 8,106	& 88.22, 0.28	& 86.51, 0.58 \\
		& $k=4$	& 12,714	& 90.00, 0.41	& 88.18, 0.28 \\
		& $k=8$	& 21,930	& 91.22, 0.25	& 88.88, 0.19 \\
\bottomrule
\end{tabular}\label{tab_fmnist_sma_conv}
\end{table}

\begin{table}[htbp]
\centering
\caption{Results on Fashion-MNIST with two hidden layers: 784--64--64--10.}
\medskip
\begin{tabular}{lrrr}
\toprule
Model 	& Params	& Train acc 	& Test acc \\
\midrule
random	& 650		& 72.16, 1.19	& 71.04, 1.24 \\
\hline
$k=2$	& 3,450	& 83.15, 0.47	& 81.52, 0.50 \\
$k=4$	& 5,146	& 84.43, 0.56	& 82.88, 0.27 \\
$k=8$	& 8,538	& 86.20, 0.57	& 84.27, 0.56 \\
$k=16$	& 15,322	& 86.99, 0.83	& 84.50, 1.00 \\
$k=32$	& 28,890	& 88.22, 0.16	& 85.99, 0.19 \\
$k=48$	& 42,458	& 89.33, 0.32	& 86.38, 0.24 \\
\hline
full 				& 55,050	& 89.75, 0.25	& 87.40, 0.27 \\
\bottomrule
\end{tabular}\label{tab_fmnist_lar_noconv}
\end{table}

\begin{table}[htbp]
\centering
\caption{Results on Fashion-MNIST with one hidden layer and binary half-layer weights.}
\medskip
\begin{tabular}{lrrr}
\toprule
Model 			& Params	& Train acc 	& Test acc \\
\midrule
random			& 650	& 76.73, 0.58	& 75.86, 0.72 \\
\hline
$k=2$	& 3,130	& 77.16, 1.77	& 76.22, 1.38 \\
$k=4$	& 4,698	& 81.69, 1.11	& 80.00, 0.97 \\
$k=8$	& 7,834	& 82.40, 0.64	& 82.85, 0.64 \\
$k=16$	& 14,106	& 85.13, 1.09	& 83.25, 1.15 \\
\bottomrule
\end{tabular}\label{tab_fmnist_bin}
\end{table}

\newpage
\section{Results on CIFAR-10}

\begin{table}[htbp]
\caption{Results on CIFAR-10.}
\medskip
\label{tab:cifar_results}
\centering
\begin{tabular}{llccc}
\toprule
Conv			& FC 				& Train acc 	& Test acc 	& Params \\
\midrule
Full 			& Full 				& 70.52, 0.32	& 69.52, 0.49	& 2,102,410 \\
\midrule
Full 			& Half ($k_f=8$)		& 67.76, 0.29	& 67.16, 0.42	& 79,242	\\
Full 			& Half ($k_f=16$)		& 69.33, 0.26	& 68.15, 0.17	& 144,778	\\
Full 			& Half ($k_f=32$)		& 70.73, 0.30	& 69.50, 0.39	& 275,850	\\
Full 			& Half ($k_f=48$)		& 71.76, 0.35	& 70.41, 0.44	& 406,922 	\\
\midrule
Half \\
($k_c=8$)	& Half ($k_f=32$)		& 63.98, 0.78	& 63.12, 0.69	 & 274,157 \\
\bottomrule
\end{tabular}
\end{table}



\end{document}